\documentclass[runningheads]{llncs}
\usepackage[utf8]{inputenc}
\usepackage{amsmath}
\usepackage{mathtools}
\usepackage{amssymb}
\usepackage[hyphens]{url}
\usepackage{hyperref}
\usepackage[final]{microtype}
\usepackage{tikz}
\usepackage{pgfplots}
\usepackage{adjustbox}
\usepackage{multirow}
\usetikzlibrary{calc}
\pgfplotsset{compat=1.14}
\usepgfplotslibrary{colorbrewer}
\usepackage{color}

\DeclareMathOperator*{\maximize}{maximize}

\DeclareMathOperator*{\argmin}{arg\,min}

\begin{document}

\pagestyle{headings}
\mainmatter

\title{Adversarial Shape Perturbations on 3D Point Clouds}
\author{Daniel Liu\inst{1} \and Ronald Yu\inst{2} \and Hao Su\inst{2}}
\institute{Torrey Pines High School, San Diego, CA, USA\\
\email{daniel.liu02@gmail.com}\\
University of California San Diego, La Jolla, CA, USA\\
\email{\{ronaldyu@ucsd.edu, haosu@eng.ucsd.edu\}}}

\maketitle

\begin{abstract}
The importance of training robust neural network grows as 3D data is increasingly utilized in deep learning for vision tasks in robotics, drone control, and autonomous driving. One commonly used 3D data type is 3D point clouds, which describe shape information. We examine the problem of creating robust models from the perspective of the attacker, which is necessary in understanding how 3D neural networks can be exploited. We explore two categories of attacks: distributional attacks that involve imperceptible perturbations to the distribution of points, and shape attacks that involve deforming the shape represented by a point cloud. We explore three possible shape attacks for attacking 3D point cloud classification and show that some of them are able to be effective even against preprocessing steps, like the previously proposed point-removal defenses.\footnote{Source code available at \url{https://github.com/Daniel-Liu-c0deb0t/Adversarial-point-perturbations-on-3D-objects}.}
\keywords{adversarial attacks, adversarial defenses, 3D point clouds, robustness, neural networks, deep learning, PointNet}
\end{abstract}

\section{Introduction}
3D data is often used as an input for controlling autonomous driving systems, robotics, drones, and other tasks that rely on deep learning. For example, in tasks like controlling self-driving cars, 3D point cloud data from LiDAR scans, in addition to radar and image data, are being used by companies like Waymo, Uber, Lyft, and NVIDIA. Point cloud data is also produced in many other applications, including LiDAR (Light Detection and Ranging) scans, RGB-D scans (\textit{e.g.}, through Microsoft Kinect), or photogrammetry of 3D objects. However, compared to the bitmap 2D images that are well-studied in many previous works, deep learning with 3D data is still poorly understood.
More specifically, there has not been many works studying the robustness of 3D deep learning, despite its importance in safety-critical applications and systems. Since, in general, the robustness of neural networks in adverse scenarios can be measured by looking at adversarial examples (\textit{eg.}, an adversarial object that looks like a car but the neural network misclassifies it as a person), we study how to generate effective attacks specific to 3D space to evaluate the robustness of 3D deep learning models in this paper. In our experiments, we focus on the task of 3D point cloud classification.

Currently, it is quite easy to create effective adversarial attacks in both 2D and 3D space by maximizing the loss of a neural network to generate an adversarial example that fools the neural network. However, in this paper, we are interested in crafting effective adversarial perturbations that satisfy certain desirable criteria, like morphing the shape of a point cloud to generate certain adversarial features, adding certain adversarial features to an existing point cloud, or being effective even against defenses. We examine two distinct categories of attacks, the well-studied ``distributional attacks'', which seek to minimize perceptibility, and our new ``shape attacks'' that modify the surface of a 3D object's point cloud. In total, we propose three new ``shape attacks'' for exploiting shape data that is native to 3D point clouds data in generating our adversarial examples. We do extensive experimental validation of their effectiveness, and show that some of them are able to break previous point-removal defenses, which represent realistic defenses or preprocessing steps for scanned point clouds.

\section{Related Work}
\subsection{Attacks}
\cite{szegedy2013intriguing} and \cite{biggio2013evasion} first examined how adversarial perturbations on 2D images could be generated on a neural network by searching for some perturbation that causes a neural network to make a wrong prediction. Later, there were many attacks proposed in \cite{goodfellow2014explaining}, \cite{madry2017towards}, \cite{kurakin2016adversarialscale}, and \cite{kurakin2016adversarialphysical} that were based on constraining the $L_p$ norm of the perturbation, which was a measure of perceptibility. In general, their proposed optimization problems are solved through projected gradient descent, in what is sometimes known as the ``iterative gradient attack''. There are also many further improvements by \cite{papernot2016limitations}, \cite{moosavi2016deepfool}, \cite{dong2018boosting}, \cite{brown2017adversarial}, \cite{moosavi2017universal}, and \cite{carlini2017towards} on perturbing 2D images. \cite{biggio2018wild} provides an overview of threat models and 2D attacks.

Only very recently has adversarial attacks been examined in 3D space. For 3D point clouds, \cite{xiang2018generating}, \cite{yang2019adversarial}, and \cite{liu2019extending} all proposed point perturbations based on shifting points using ideas similar to attacks on 2D images. An adversarial attack that shifted the distribution of points on a 3D object's surface was proposed in~\cite{liu2019extending} as an extremely imperceptible attack, by using projected gradient descent.

\cite{xiang2018generating} also proposed an attack that generates clusters of points with different shapes. For this attack, they optimize an objective function that constrains the distance between the large, perceptible generated point clusters and the main benign point cloud.

\cite{zheng2018learning} and \cite{wicker2019robustness} both proposed saliency-based techniques for adversarially removing points. This is different from our attacks in this paper since we only consider shifting points.
Adversarial attacks on a simulated LiDAR system were proposed in~\cite{cao2019adversarial} by making the scanning operation that converts an object's surface into a point cloud differentiable. The resulting adversarial examples generated from a basic box object are large, noticeable shape deformations. We adopt a similar approach in this paper, except our shape deformation are directly performed on raw point clouds and we directly attack point cloud neural networks.

\subsection{Defenses}
A popular defense for 2D image space is adversarial training, where a neural network is trained with adversarial examples~\cite{goodfellow2014explaining}. This is done by adding the loss incurred by adversarial examples to the objective function that is optimized throughout training. Another defensive technique is defensive distillation~\cite{papernot2016distillation}. However, so far, it is still quite easy to attack and bypass defenses in the 2D image domain. Later work has shown that a neural network can be trained to be provably robust~\cite{wong2017provable} to some degree. For neural networks that classify 3D point clouds, removing points was shown to be more effective than adversarial training. In particular, \cite{liu2019extending} and \cite{zhou2018deflecting} both examined removing outlier points, and \cite{liu2019extending} also examined removing salient points. These defenses were shown to be effective even when the adversarial perturbations were small and imperceptible.

\subsection{3D Deep Learning}
3D deep learning has recently experienced exponential growth. Many network architectures were proposed for different tasks in 3D deep learning. We specifically examine the case of 3D point cloud classification, using the PointNet~\cite{qi2017pointnet}, PointNet++~\cite{qi2017pointnetplusplus}, and DGCNN~\cite{dgcnn} architectures. Other work on this task include \cite{qi2017frustum} and \cite{deng2018ppf}. Point cloud networks were previously shown to be robust against small random perturbations and random point dropout. There is also work on handling voxel (quantized point cloud) data~(\cite{wang2017cnn}, \cite{wu2015shapenet}), though this is not examined in this paper.

\section{Setting}
We have a neural network model $f_\theta: \mathbb{R}^{N \times 3} \mapsto \mathbb{R}^M$, which solves the 3D point cloud classification task by predicting probability vectors for each output class. It takes in $x$, a set of $N$ 3D points, and outputs a vector $f_\theta(x)$ of length $M$ that contains the probability of the input $x$ being each output possible class. The model $f_\theta$ is trained by adjusting its parameters $\theta$ to minimize the cross entropy loss, which is denoted by $J(f_\theta(x), y)$, for each sample $x$ and its corresponding one-hot label $y$.

\subsubsection{Threat model.} We assume that the attacker has full access to the architecture and parameters $\theta$ of a neural network $f_\theta$ (white-box threat model). The attacker also has access to the unstructured 3D point clouds, which they can change before feeding them to the neural network. Therefore, we wish to construct attacks that are purely based on unstructured 3D point sets, without any extra shape or normal information on the structure of the point set. Our experiments occur in a \textit{digital} setting, and we leave experimental evaluations of our attacks on \textit{physical} objects as future work. We will focus on generating \textit{untargeted} attacks, where the attacker is attempting to force a neural network into misclassifying an object of a certain class into a different class. However, our algorithms can be easily extended to targeted attacks.

In our experiments, we also evaluate defensive techniques. After we generate adversarial attacks on a clean/benign point cloud $x$ by perturbing it with $\delta \in \mathbb{R}^{N \times 3}$ to obtain $x^\ast = x + \delta$, we feed $x^\ast$ to a defense before it is classified by a neural network.
We now discuss some previously proposed defenses that we will be used in our attack evaluation. These defenses will help simulate a realistic scenario, where the scanning operation and subsequent preprocessing, like outlier removal, make generating effective attacks much more difficult.

\subsubsection{Random point removal.} We test basic random point dropout, which randomly selects and removes a set of points from $x$. We do not expect this to be a very effective defense.

\subsubsection{Outlier removal.} This defense proposed in \cite{liu2019extending} involves first calculating statistical outliers through
\begin{equation}
o[i] = \frac{1}{K} \sum_{j = 1}^K \operatorname{NN}(x^\ast[i], j), \quad \forall i \in \{1 \ldots N\}
\end{equation}
with $\operatorname{NN}(\rho, j)$ returning the $j$-th nearest neighboring point of the point $\rho$. Then, we remove points greater than $\epsilon$ standard deviations away from the average $o[i]$ across all $i$:
\begin{equation}
\begin{aligned}
x^{\ast\ast} = x^\ast \setminus \Big\{x^\ast[i] : o[i] > \frac{1}{N} \sum_{j = 1}^N o[j] + \epsilon \operatorname{stddev}(o)
\wedge i \in \{1 \ldots N\}\Big\}
\end{aligned}
\end{equation}
\subsubsection{Salient point removal.} In this defense proposed by \cite{liu2019extending}, salient points are identified and removed by first calculating the saliency
\begin{equation}
s[i] = \max_{j \in \{1 \ldots M\}} \Big|\Big|\big(\nabla_{x^\ast} f_\theta(x^\ast)[j]\big)[i]\Big|\Big|_2, \quad \forall i \in \{1 \ldots N\}
\end{equation}
and a subset of the points in $x^\ast$ with the highest saliencies are removed.

\begin{table*}[tp]
    \centering
    \caption{How our new shape attacks compare to some previous types of attacks in both 2D and 3D domains. We mainly focus on \textit{shape attacks} in this paper.}
    
    \begin{adjustbox}{width=\textwidth}
    \setlength{\tabcolsep}{0.5em}
    {\renewcommand{\arraystretch}{1.5}%
    \begin{tabular}{p{1cm}|p{12.5cm}}
        \hline
        Images & \textbf{Small perturbations for all pixels.} $L_p$ norm attacks like \cite{goodfellow2014explaining}, \cite{dong2018boosting}, \cite{szegedy2013intriguing}, \cite{carlini2017towards}, \cite{moosavi2016deepfool}, and other attacks that minimize perceptibility. Small perturbations are easily lost in real photographs or against defenses/preprocessing.\\
        \cline{2-2}
        & \textbf{Large perturbations for some pixels.} Either strong single-pixel perturbations~\cite{papernot2016limitations}, low frequency attack~\cite{guo2018low}, or large adversarial patches~\cite{brown2017adversarial} that are visually noticeable and relatively easy to apply. More likely to be effective against preprocessing and in physical domains (taking a photo of the adversarial examples) due to noticeable perturbations.\\
        \hline
        Point Clouds & \textbf{Distributional attacks.} Small perturbations or imperceptible point insertion or removal, like in \cite{liu2019extending}, \cite{zheng2018learning}, \cite{wicker2019robustness}, and \cite{xiang2018generating} (some of which are inspired by 2D $L_p$ methods). Similar to its 2D analog, small point perturbations are easily lost due preprocessing/defenses.\\
        \cline{2-2}
        & \textbf{Shape attacks.} Morphing the shape of the point cloud by changing multiple points in some focused areas of the point cloud, instead of single point perturbations. \\
        \cline{2-2}
        & \textbf{Adding disjoint clusters.} The attack in~\cite{xiang2018generating} works by adding many clusters of smaller point clouds to an existing point cloud. This is extremely noticeable (similar to our shape attacks), and generates disjoint objects.\\
        \hline
    \end{tabular}}
    \end{adjustbox}
    
    \label{table:prev_work}
\end{table*}

\section{Attack Types}
In this paper, we present two classes of attacks that represent contradictory attack objectives: \textit{distributional} and \textit{shape} attacks. We summarize the benefits and drawbacks of each category of attacks (along with their 2D counterparts) in Table~\ref{table:prev_work}. We elaborate on the distributional and shape attacks below:

\subsubsection{Distributional attacks.} In this paper, we will mainly focus on distributional attacks that attempt to perturb every point by a small, imperceptible amount (for human eyes) on or very near a 3D object. Usually, the objective functions for these distributional attacks will merely involve minimizing perceptibility through some distance function while maximizing the loss of a neural network. However, due to the spread out, low magnitude perturbations generated in these attacks, we show that the adversarial examples will be brittle against some existing defenses and we believe that they are also difficult to realize on real life objects due to the easily ignored perturbations.

\subsubsection{Shape attacks.} These attacks create larger perturbations at a select few locations to create more noticeable continuous shape deformations without outliers (as opposed to outlying points created with previous distributional attacks in \cite{liu2019extending} and \cite{xiang2018generating}) through the use of resampling, etc. We show that this is robust against previous point-removal defenses due to the continuous, outlier-free changes. We believe that the larger perturbations on real objects are also more likely to be picked up by scanning operations, and the clear, focused deformations in a few areas may be easier to physically construct on real life objects than small perturbations for every point in a point cloud. These attacks may make use of more complex objective functions than just minimizing perceptibility.

\subsubsection{Constructing physical adversarial examples.} Although we do not discuss this in detail in this paper, we note that these shape attacks can be realized through 3D printing with triangulated point clouds~(\cite{tsai2020robust}, \cite{edelsbrunner1983shape}, \cite{bernardini1999ball}), forceful shape deformations (\textit{eg.}, bending, breaking, etc.), or adding new features with flimsy/cheap materials. These can be done by using generated adversarial shape attacks as a guide to modify specific regions of a physical object.

\section{Distributional Attacks}
We define a distributional attack as any attack that optimizes for imperceptible point perturbations:
\begin{equation}
\maximize_\delta \quad J(f_\theta(x + \delta), y) - \lambda \mathcal{D}(x + \delta, x)
\end{equation}
where $\mathcal{D}$ is some metric for measuring the perceptibility of an adversarial attack. A variant of this objective is constraining $\mathcal{D}(x + \delta, x) < \epsilon$, for some threshold $\epsilon$. These definitions will include attacks from previous works like \cite{xiang2018generating} and \cite{liu2019extending} (for example, the iterative gradient $L_p$ attack or projected gradient descent attacks in 3D). Here, we give a basic overview of the previous Chamfer~\cite{xiang2018generating} and gradient projection~\cite{liu2019extending} attacks that were previously proposed.

\subsection{Chamfer Attack}
There are a few ways to measure the distance (and thus, perceptibility) between two point sets. One way is the Chamfer distance, which is given by
\begin{equation}
\mathcal{C}(A, B) = \frac{1}{|A|}\sum_{a \in A} \min_{b \in B} ||a - b||_2
\end{equation}
for two sets $A$ and $B$. With this distance and the $L_2$ norm to discourage large perturbations, the following optimization problem can be solved using an algorithm like Adam~\cite{kingma2014adam} to generate a distributional adversarial example:
\begin{equation}
\maximize_\delta \quad J(f_\theta(x + \delta'), y) - \lambda \big(\mathcal{C}(x + \delta', x) + \alpha ||\delta'||_2\big)
\end{equation}
where the $L_\infty$ norm of the perturbation $\delta'$ is hard bounded using $\delta' = \tanh{\delta}$ (note that each point cloud is scaled to fit in a unit sphere), so the perturbations cannot become arbitrarily large. The generated adversarial example is $x^\ast = x + \delta'$. We binary search for $\lambda$ and set a value $\alpha$ for balancing the two distance functions. This is one of the proposed attacks in \cite{xiang2018generating}, and in this paper we will refer to it as the \textit{Chamfer attack}. In general, the important of using the Chamfer metric to measure perceptiblity is evident when compared to using a standard $L_p$ norm to measure perceptibility~\cite{liu2019extending}, as $L_p$ norms do not encourage perturbations that result in point shifts across the surface of an object (note that the points $x$ lies on the surface of an object).

\subsection{Gradient Projection}
For our variant of the gradient projection attack~\cite{liu2019extending} that we evaluate, we use a very similar metric to Chamfer distance: the Hausdorff distance
\begin{equation}
\mathcal{H}(A, B) = \max_{a \in A} \min_{b \in B} ||a - b||_2
\end{equation}
Although this metric was also used in previous works like \cite{xiang2018generating}, our new attack variant involves measuring the amount of perturbation by comparing the perturbed point cloud $x^\ast$ to the benign surface $S$ of a point cloud, rather than the benign point cloud $x$. This is less sensitive to the density of points, and imperceptible point shifts along the surface of the object do not increase the Hausdorff distance. Note that in practice, the benign surface $S$ is represented through a triangular mesh $t$ with $T$ triangles. This triangular mesh can be obtained through an algorithm like the alpha shape algorithm~\cite{edelsbrunner1983shape} using the 3D Delaunay triangulation~\cite{lee1980two} of each point cloud (other triangulation algorithms like \cite{bernardini1999ball} also work).

If we structure our optimization problem for constructing adversarial point clouds to maximize $J$ while bounding the Hausdorff distance with an easy-to-tune parameter $\tau$, then we can craft adversarial perturbations that are imperceptible. The problem can be formulated as
\begin{equation}
\begin{aligned}
&\maximize_\delta \quad J(f_\theta(x + \delta), y),
&\text{subject to} \quad \mathcal{H}(x + \delta, S) \leq \tau
\end{aligned}
\end{equation}
This formulation is similar to the optimization problem solved through \cite{madry2017towards}'s projected gradient descent, but with a different method for measuring perceptibility. Therefore, it can be solved the same way, using $n$ steps of gradient descent (unfortunately, momentum cannot be used) constrained under the $L_2$ norm by $\frac{\epsilon}{n}$ (for some hyperparameter $\epsilon$) and then projecting onto the set $\{a : \mathcal{H}(\{\rho\}, S) \leq \tau \wedge \rho \in \mathbb{R}^3\}$ for a surface $S$ using a variant of the basic point-to-triangle projection, since $S$ is represented by a set of triangles $t$. In our implementation of this attack, we use a VP-tree~\cite{yianilos1993data} to organize the 3D triangles, which are represented as points, for faster projections by pruning with a calculated query radius in the tree. Note that this algorithm is a generalization of the gradient projection attack proposed in \cite{liu2019extending}, which is just a special case where $\tau = 0$.

\section{Shape Attacks}
The general idea behind each of our proposed shape attacks is that we try to use point perturbations that are known to be successful in the distributional attacks, and then somehow ``connecting'' these points with the main point cloud object to create shape deformations, so that the resulting adversarial point cloud is a \textit{single} connected object without any outliers. This is partially inspired by low-frequency attacks on 2D images~\cite{guo2018low}, which generates smooth perturbations over multiple pixels, instead of noisy perturbations for each pixel. In our attacks, we enforce the criteria that the \textit{density} of the sampled points to be as uniform as possible on $S^\ast$, the surface of an adversarial object, through methods like resampling.

\subsection{Perturbation Resampling}
Consider an adversarial object's surface that is represented by the set $S^\ast$. We want to ensure that the point distribution of a point cloud $x^\ast \subset S^\ast$ is approximately uniform density on this surface by maximizing the following objective:
\begin{equation}
\begin{aligned}
&\maximize_{x^\ast} &&\min_{i, j \in \{1 \ldots N\}, i \neq j} ||x^\ast[i] - x^\ast[j]||_2,
&\text{subject to} &&x^\ast \subset S^\ast
\end{aligned}
\end{equation}
in addition to maximizing the loss, while constraining the perceptibility. The easiest method to accomplish this is through the use of a point resampling algorithm, like farthest point sampling, on the estimated shape $S^\ast$ (which is obtained using something like the alpha shape algorithm to get a triangulation of the point cloud) after perturbing the point cloud with gradient descent. This simple idea of adding resampling into the optimization process was also examined in a parallel work~\cite{tsai2020robust}, which also proposed the usage of sampling points on the surface of the estimated shape in the attack.

Overall, the attack procedure can be summarized as the following: during each iteration of the perturbation resampling attack, we first execute one step of gradient descent (without momentum, due to the resampling operation) constrained by the $L_2$ norm and perturb the point cloud $x$ to get $x^\ast$. Then, we approximate the shape of this perturbed point cloud by computing the alpha shape to get $S^\ast$. Finally, we use farthest point sampling to resample the $\kappa$ points with the lowest saliencies (which is defined as the $L_2$ norm of the gradient of the loss function wrt each point) onto $S^\ast$. The $N - \kappa$ points with higher saliencies are automatically considered ``picked'' from sampling.
Intuitively, we are essentially perturbing the underlying shape of the 3D point cloud, while ensuring that points are evenly sampled on the surface of that shape when the surface is stretched due to the perturbations.

\subsection{Adversarial Sticks}
We can also attempt to create new features on the mesh instead of making small, incremental deformities through the perturbation resampling attack. \cite{xiang2018generating} explores this idea by adding clusters of points and even 3D objects from other classes to attack a benign 3D object. Their attack crafts adversarial examples that can contain many disjoint point clouds. We think that this is unrealistic for a classification task on a single point cloud object, so we attempt to add a very simple new feature (sticks, or line segments) onto a point cloud, where the sticks must originate from the surface of the benign 3D object. This shows that adding only a few new features is an effective method for generating adversarial point clouds.

\begin{figure*}[thp]
    \centering
    \includegraphics[width=0.5\linewidth,trim={9.2cm 0 0 0},clip]{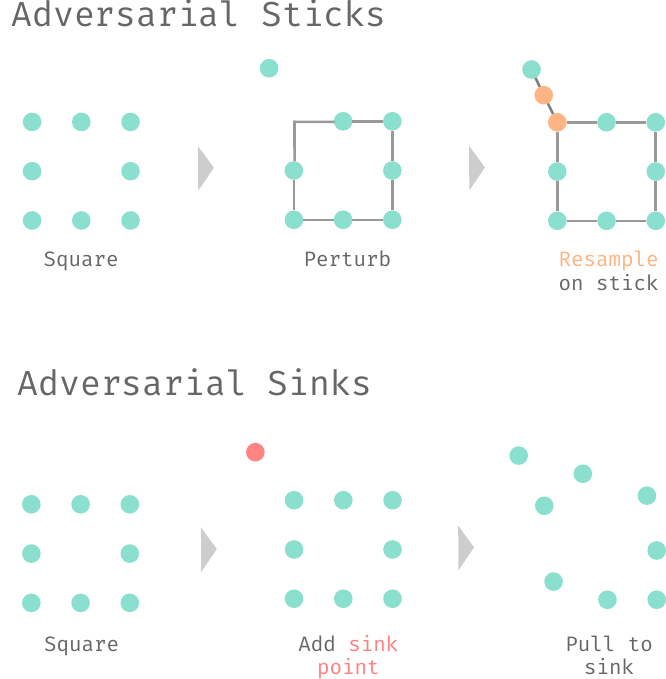}
    
    \caption{2D point cloud of a square, demonstrating the general idea for our adversarial sticks and adversarial sinks attacks. 2D line segments in the image correspond to an estimated triangular mesh in 3D.}
    \label{fig:attacks_2d}
\end{figure*}

A straightforward formulation of the problem results in optimizing over the position vectors $\rho \in \mathbb{R}^{\sigma \times 3}$ and the orientation and length vectors $\delta \in \mathbb{R}^{\sigma \times 3}$ for generating $\sigma$ sticks:
\begin{equation}
\begin{aligned}
&\maximize_{\rho, \delta} &&J\big(f_\theta \big(x \cup \mathcal{S}(\rho, \rho + \delta', \kappa(\delta'))\big), y\big) - \lambda \big(\mathcal{C}(\rho + \delta', x) + \alpha ||\delta'||_2\big),\\
&\text{subject to} &&\rho \subset S
\end{aligned}
\end{equation}
The adversarial examples will be $x^\ast = x \cup \mathcal{S}(\rho, \rho + \delta', \kappa(\delta'))$. Note that $\delta' = \frac{1}{2} \tanh \delta$, and $\mathcal{S}(\rho, \rho + \delta', \kappa(\delta'))$ samples $\kappa(\delta')$ points across the line segments defined by the points $\rho[i]$ and $\rho[i] + \delta'[i]$, $\forall i \in \{1 \ldots \sigma\}$. In practice, we use farthest point sampling to pick which points to keep unperturbed and resample the rest onto the adversarial sticks, instead of adding new points. As the lengths of the generated adversarial sticks can vary, we calculate
\begin{equation}
\kappa(\delta') = \bigg\lfloor\frac{\sum_{i = 1}^\sigma ||\delta'[i]||_2}{\mu'}\bigg\rfloor, \quad\text{where} \quad\mu' = \frac{1}{\mu N} \sum_{i = 1}^N \min_{j \in \{1 \ldots N\} \setminus \{i\}} \big|\big|x[i] - x[j]\big|\big|_2
\end{equation}
and assign points to sticks based on the lengths of the sticks. The hyperparameter $\mu$ can be adjusted to change the relative density of the sampled points on the adversarial sticks and the number of sampled points. Solving this optimization problem is difficult due to a number of factors. First, the sampling function $\mathcal{S}$ is not differentiable, which means that the objective function cannot be directly maximized. Second, we need to somehow constrain the position of each stick onto the surface of the benign point cloud $x$ in each step of the attack.

To solve these problems, we can approximate the solution to the optimization problem by choosing and perturbing $\sigma$ points with the highest saliencies ($L_2$ norm of the gradient of the loss function with respect to each input point) instead of orienting sticks. Then, this can be optimized using Adam~\cite{kingma2014adam}, similar to the Chamfer attack, except a mask is used so only $\sigma$ points are perturbed. We also hard bound the perturbations with $\frac{1}{2} \tanh$. After perturbing those points, we connect them to the surface $S$ of a point cloud, so we get $\rho = \mathcal{P}(x')$ and $\delta' = x' - \mathcal{P}(x')$, if the $x' \in \mathbb{R}^{\sigma \times 3}$ is the Chamfer adversarial example. $\mathcal{P}$ is a function that projects each point onto the benign surface $S$ (which is represented as a set of triangles $t$ in practice) of a point cloud:
\begin{equation}
\begin{aligned}
\mathcal{P}(x') &= \bigcup_{i = 1}^N \Big\{\argmin_{p \in \operatorname{proj}(x'[i])} ||x'[i] - p||_2\Big\},\quad\text{where}\\
\operatorname{proj}(p) &= \bigcup_{j = 1}^{T} \left\{\operatorname{clip\_to\_tri}\Big(p - \big((p - t[j][0]) \cdot n[j]\big)n[j],\ t[j]\Big)\right\}
\end{aligned}
\end{equation}
Note that $n[j]$ is the unit normal vector of the $j$-th triangle and $\operatorname{clip\_to\_tri}$ clips an orthogonally projected point to be inside a triangle by examining each edge of the triangle. This projection can be done efficiently using a VP-tree~\cite{yianilos1993data}.

\subsection{Adversarial Sinks}
Due to resampling and projection operations used in the adversarial sinks attack that make optimization difficult, the natural question is whether it is possible to create new shape features without those operations. Indeed, it is possible to deform the shape of the benign point cloud to generate an adversarial example without inferring its surface. We can guide this deformation through sink points, which pull points in the point cloud in a local spatial region towards them. This means that there will not be outlier points after the attack. In practice, this should lead to smooth shape deformations. The reason for guiding the deformation process with a few points is because we know that perturbing a select few points is effective against some point cloud neural networks due to their max pooling operation that selects a subset of critical points in their prediction process~(\cite{liu2019extending}, \cite{qi2017pointnet}).

Given a set of $\sigma$ sink points $s \in \mathbb{R}^{\sigma \times 3}$ that are initialized with $s_0 \in \mathbb{R}^{\sigma \times 3}$, we can deform the point cloud $x$ to create $x^\ast$ with
\begin{equation}
\begin{aligned}
x^\ast[i] &= x[i]
+ \tanh\Big(\sum_{j = 1}^{\sigma} (s[j] - x[i]) \phi(||s_0[j] - x[i]||_2)\Big),
\quad\forall i \in \{1 \ldots N\}
\end{aligned}
\label{eq:sinks}
\end{equation}
where $\phi$ represents a radial basis function that decreases the influence of each sink point by distance. This means that points in the point cloud that are far away from a sink point's initial position will be influenced less by that sink point as it moves. We use the Gaussian radial basis function, which is defined as
\begin{equation}
\begin{aligned}
&\phi(r) = e^{-\big(\frac{r}{\mu'}\big)^2}, &\text{where}\quad
\mu' = \frac{\mu}{N} \sum_{i = 1}^N \min_{j \in \{1 \ldots N\} \setminus \{i\}} \big|\big|x[i] - x[j]\big|\big|_2
\end{aligned}
\end{equation}
The hyperparameter $\mu$ can be tuned to control how the influence of each sink point falls off as distance increases. The influence of each sink point is also scaled by the average distance between point in $x$ and its nearest neighboring point, so it is invariant to the scale of the point cloud $x$. We choose the Gaussian radial basis function because it is differentiable everywhere and it outputs a value between 0 and 1, which represents a percentage of the distance between each point cloud point and sink point. In essence, Equation~\ref{eq:sinks} represents shifting each point in the point cloud by the weighted sum of the vectors from that point to each sink point.

To calculate $s_0$, we would perturb a subset of the points in $x$ with the highest saliencies by a small amount based on their saliencies, and then use farthest point sampling to chose $\sigma$ sink points. This maximizes the distance between the sink points to perturb different parts of the point cloud and initializes sink points in locations that may lead to effective perturbations.

To craft an adversarial example, we solve the following optimization problem:
\begin{equation}
\begin{aligned}
\maximize_{s} &\quad J\big(f_\theta (x^\ast), y\big) - \lambda \Big(||x^\ast - x||_2 + \alpha \max_{i \in \{1 \ldots \sigma\}} \big|\big|s[i] - s_0[i]\big|\big|_2\\
&\quad - \beta \min_{i, j \in \{1 \ldots \sigma\}, i \neq j} \big|\big|s[i] - s[j]\big|\big|_2\Big)
\end{aligned}
\end{equation}
where $x^\ast$ is constructed according to Equation~\ref{eq:sinks}. Solving this problem means maximizing the loss and the distance between sinks (to avoid perturbations interfering with each other) by shifting our sink points $s$, while minimizing the $L_2$ norm of the perturbation (similar to the Chamfer attack, $\tanh$ is used to hard bound the $L_\infty$ norm of the perturbations) and the distance between the sink point and its corresponding initial sink point. $\lambda$ can be adjusted using binary search.
Since this objective function is fully differentiable, we can directly apply a gradient-based optimization algorithm like Adam~\cite{kingma2014adam} to maximize our objective function.

\begin{figure*}[thp]
    \begin{adjustbox}{width=0.9\textwidth,center}
    \includegraphics{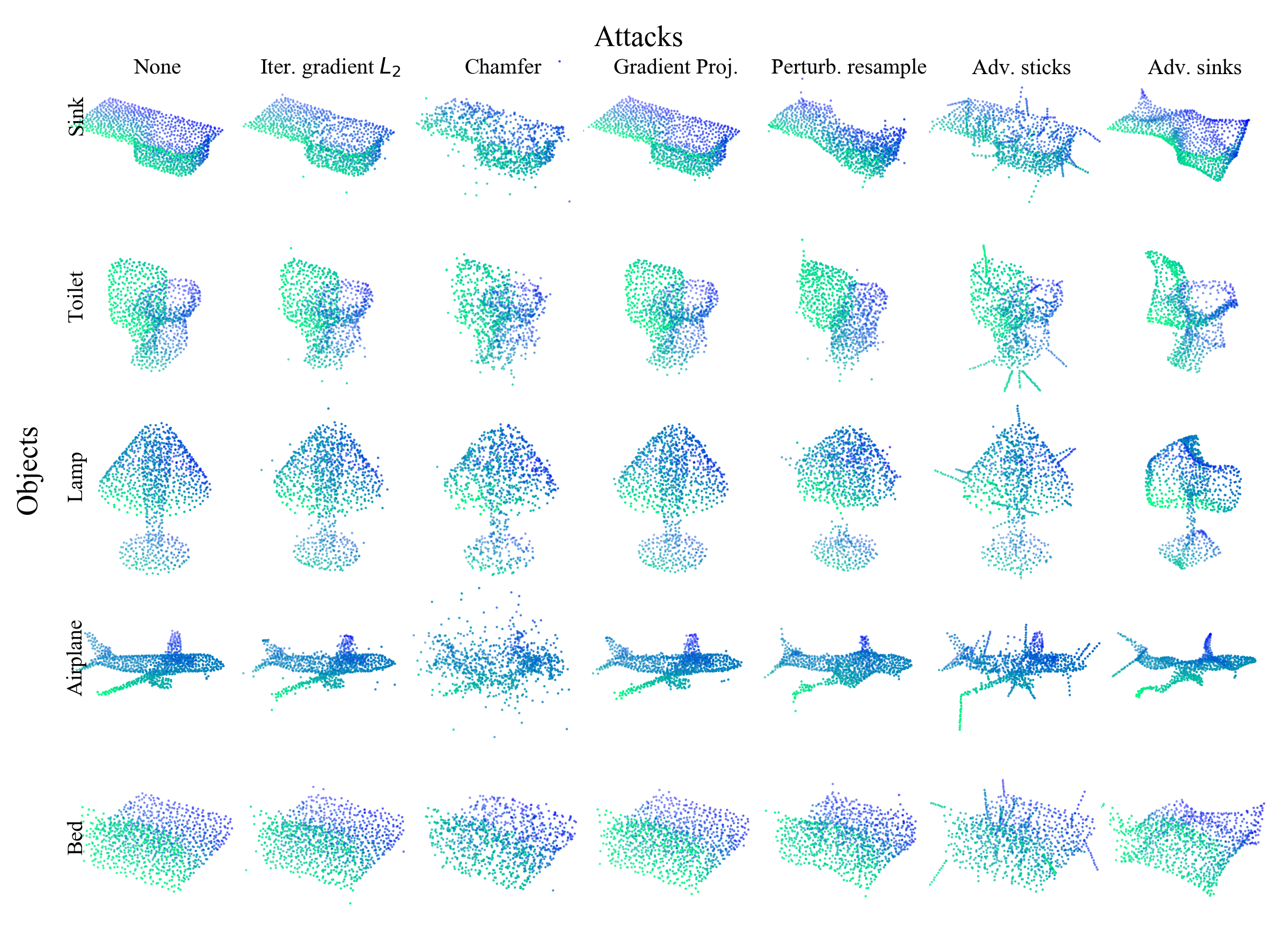}
    \end{adjustbox}
    
    \caption{Visualizations of adversarial examples that are generated on the PointNet architecture, for five different ModelNet40 classes. Notice that the Chamfer, gradient projection, and iterative gradient $L_2$ attacks produce noisy point clouds, while the perturbation resampling, adversarial sticks, and adversarial sinks attacks change the shape of the point cloud objects. Note that attacks with bounded perturbations (iterative gradient $L_2$, gradient projection, and perturbation resampling) cannot automatically generate larger perturbations to obtain more successful adversarial examples. There are some areas that are noticeably missing points in the perturbation resampling adversarial examples, which is due to limitations in the triangulation procedure on a fixed number of points. Other triangulation methods may produce better point clouds, but we leave this as an open question that we do not explore further.}
    \label{fig:object_attack}
\end{figure*}

\begin{table*}[tp]
    \centering
    \caption{The success rates of untargeted adversarial attacks against different defenses on the PointNet, PointNet++, and DGCNN architectures. The highest attack success rate for each neural network and defense combination is bolded. The last three columns are our new shape attacks.}
    
    \begin{adjustbox}{width=\textwidth,center}
    \setlength{\tabcolsep}{0.7em}
    {\renewcommand{\arraystretch}{1.2}%
    \begin{tabular}{cl|c|ccc|ccc}
        \hline
        & &\multicolumn{7}{c}{Attacks}\\
        & Defenses & None & Iter. grad. $L_2$ & Chamfer & Grad. proj. & Perturb. resample & Adv. sticks & Adv. sinks\\
        \hline
        
        \multirow{4}{*}{\rotatebox[origin=c]{90}{PointNet}} & None & 0.0\% & 99.1\% & \textbf{100.0\%} & 89.8\% & 97.5\% & 93.8\% & 99.9\%\\
        & Random remove & 0.8\% & 98.4\% & \textbf{98.7\%} & 86.3\% & 96.4\% & 93.2\% & 95.3\%\\
        & Remove outliers & 5.6\% & 56.9\% & 49.3\% & 46.9\% & 81.6\% & \textbf{86.8\%} & 84.7\%\\
        & Remove salient & 7.7\% & 51.8\% & 55.5\% & 41.6\% & 82.8\% & \textbf{89.5\%} & 88.5\%\\
        
        \hline
        
        \multirow{4}{*}{\rotatebox[origin=c]{90}{PointNet++}} & None & 0.0\% & 100.0\% & 97.7\% & 98.8\% & 98.4\% & 85.4\% & \textbf{99.8\%}\\
        & Random remove & 3.6\% & \textbf{96.2\%} & 96.0\% & 75.9\% & 95.2\% & 83.4\% & 83.1\%\\
        & Remove outliers & 11.0\% & 44.4\% & 62.9\% & 29.6\% & 86.7\% & 79.1\% & \textbf{87.3\%}\\
        & Remove salient & 10.2\% & 46.6\% & 82.1\% & 26.8\% & 81.0\% & 79.8\% & \textbf{84.5\%}\\
        
        \hline
        
        \multirow{4}{*}{\rotatebox[origin=c]{90}{DGCNN}} & None & 0.0\% & 42.0\% & 99.6\% & 34.5\% & 76.0\% & 83.2\% & \textbf{100.0\%}\\
        & Random remove & 5.6\% & 32.4\% & 78.1\% & 22.1\% & 72.8\% & \textbf{84.3\%} & 77.5\%\\
        & Remove outliers & 10.5\% & 22.6\% & 46.8\% & 18.4\% & 69.1\% & 66.9\% & \textbf{85.5\%}\\
        & Remove salient & 11.9\% & 24.1\% & 51.3\% & 18.0\% & 65.8\% & \textbf{81.3\%} & 73.0\%\\
        
        \hline
    \end{tabular}}
    \end{adjustbox}
    
    \label{table:pointnet}
\end{table*}

\section{Results}
\subsection{Setup}
\subsubsection{Models and dataset.} We train PointNet~\cite{qi2017pointnet}, PointNet++~\cite{qi2017pointnetplusplus}, and DGCNN~\cite{dgcnn} with default hyperparameters, but we use a slightly lowered batch size for PointNet++ due to memory constraints. The neural networks are trained on the training split of the ModelNet40 dataset~\cite{wu2015shapenet}. The test split of the dataset is used for evaluating our adversarial attacks. We use point clouds of size $N = 1,024$ sampled with uniform density from 3D triangular meshes in the ModelNet40 dataset.

\subsubsection{Attacks.} For the iterative gradient $L_2$ attack, we use $\epsilon = 2$ and $n = 100$ iterations. For the Chamfer attack, we run the Adam optimizer ($\beta_1 = 0.9$ and $\beta_2 = 0.999$) for $n = 20$ iterations with a learning rate of $\eta = 0.1$, we balance the two distance functions with $\alpha = 0.002$, and we binary search for $\lambda$. For the gradient projection attack, we use $\epsilon = 1$, $\tau = 0.05$, and $n = 20$ iterations (it is relatively more time consuming to run than other attacks). For perturbation resampling, we use $\epsilon = 2$, we resample $\kappa = 500$ points, and we run the attack for $n = 100$ iterations. For adversarial sticks, we use the Adam optimizer ($\beta_1 = 0.9$ and $\beta_2 = 0.999$) for $n = 20$ iterations, with $\eta = 0.1$ and $\alpha = 0.01$ (we binary search for $\lambda$). We add $\sigma = 100$ sticks, and we resample points with $\mu = 2$. For adversarial sinks, we use a learning rate of $\eta = 0.1$ for the Adam optimizer ($\beta_1 = 0.9$ and $\beta_2 = 0.999$), we scale the strength of each sink point by $\mu = 7$, we use $\alpha = 5$ and $\beta = 1$, and we run the attack for $n = 20$ iterations with $\sigma = 30$ sink points. The hyperparameters were determined through rudimentary grid search, and we will show the effect of changing some of them in our experiments. Additionally, we use less iterations for the attacks that use the Adam optimizer instead of vanilla gradient descent because Adam is much more efficient. Note that directly comparing the attacks that use binary search with those that do not is unfair.

\subsubsection{Defenses.} For random point removal, we drop out 200 random points. For removing outliers, we calculate the average distance from each point to its 10 nearest neighbors, and remove points with an calculated average distance greater than one standard deviation from the average of the average distances across all points. For removing salient points, we remove the 200 points with the highest saliencies. We do not test adversarial training because it is not attack agnostic and it was found to perform worse than the point removal defenses~\cite{liu2019extending}.

\subsection{Adversarial Attacks}
In Table~\ref{table:pointnet}, we show the success rates of our attacks on the 2,000+ correctly classified objects from the ModelNet40 dataset, against the PointNet, PointNet++, and DGCNN architectures with different defenses. Adversarial examples are visualized in Figure~\ref{fig:object_attack}.

From Table~\ref{table:pointnet}, we see that overall, our new attacks are quite successful against the defended PointNet, PointNet++, and DGCNN models, but they are slightly less effective than the traditional iterative gradient $L_2$ attack when there are no strong defenses. This is due to the extra processes like projection, resampling, and limiting the perturbations to a few select areas, which are used to meet the desired perceptibility or shape deformation criteria. Adversarial sinks seems to be the most successful, as expected, since its objective function easier to optimize. The effectiveness of the shape attacks against the defenses can be explained through the harder to remove/ignore deformation features, the inability for the defense to rely on unperturbed points with no gradient flow (due to the max pooling operation) for a correct prediction, and the inability for the defense to identify and remove adversarial outliers.
We note that the random point removal defenses is very weak compared to the other defenses.

\pgfplotsset{
    cycle list/Dark2,
    cycle multiindex* list={
        mark list*\nextlist
        Dark2\nextlist
    }
}
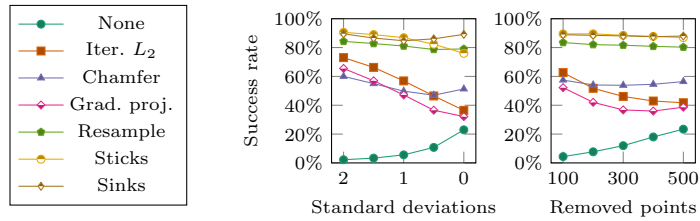
\begin{figure}[tp]
    \centering
    \hspace{7.3em}
    \begin{tikzpicture}
    \begin{axis}[
        name=left,
        xlabel=\scriptsize{Standard deviations},
        ylabel=\scriptsize{Success rate},
        ymin=0,
        ymax=1,
        width=3.5cm,
        height=3.5cm,
        x dir=reverse,
        legend style={
            overlay,
            at={(-1,0.4)},
            anchor=east,
            font=\scriptsize},
        yticklabel={\scriptsize\pgfmathparse{\tick*100}\pgfmathprintnumber{\pgfmathresult}\%},
        xticklabel={\scriptsize\pgfmathprintnumber{\tick}}
    ]
        \addplot+[mark size=1.5pt] table [x=param,y=none,col sep=comma] {outlier_defense.csv};
        \addplot+[mark size=1.5pt] table [x=param,y=iter_l2,col sep=comma] {outlier_defense.csv};
        \addplot+[mark size=1.5pt] table [x=param,y=chamfer,col sep=comma] {outlier_defense.csv};
        \addplot+[mark size=1.5pt] table [x=param,y=distrib,col sep=comma] {outlier_defense.csv};
        \addplot+[mark size=1.5pt] table [x=param,y=resample,col sep=comma] {outlier_defense.csv};
        \addplot+[mark size=1.5pt] table [x=param,y=sticks,col sep=comma] {outlier_defense.csv};
        \addplot+[mark size=1.5pt] table [x=param,y=sinks,col sep=comma] {outlier_defense.csv};
        \legend{None,Iter. $L_2$,Chamfer,Grad. proj.,Resample,Sticks,Sinks}
    \end{axis}
    \begin{axis}[
        xlabel=\scriptsize{Removed points},
        ymin=0,
        ymax=1,
        width=3.5cm,
        height=3.5cm,
        at={($(left.north east)+(1cm,0)$)},
        anchor=north west,
        yticklabel={\scriptsize\pgfmathparse{\tick*100}\pgfmathprintnumber{\pgfmathresult}\%},
        xtick={100,300,500},
        xticklabels={\scriptsize{100},\scriptsize{300},\scriptsize{500}}
    ]
        \addplot+[mark size=1.5pt] table [x=param,y=none,col sep=comma] {saliency_defense.csv};
        \addplot+[mark size=1.5pt] table [x=param,y=iter_l2,col sep=comma] {saliency_defense.csv};
        \addplot+[mark size=1.5pt] table [x=param,y=chamfer,col sep=comma] {saliency_defense.csv};
        \addplot+[mark size=1.5pt] table [x=param,y=distrib,col sep=comma] {saliency_defense.csv};
        \addplot+[mark size=1.5pt] table [x=param,y=resample,col sep=comma] {saliency_defense.csv};
        \addplot+[mark size=1.5pt] table [x=param,y=sticks,col sep=comma] {saliency_defense.csv};
        \addplot+[mark size=1.5pt] table [x=param,y=sinks,col sep=comma] {saliency_defense.csv};
    \end{axis}
    \end{tikzpicture}
    
    \caption{The success rates of adversarial attacks against the PointNet architecture as we increase the number of points removed with the outlier removal (left) and the salient point removal (right) defenses. For the outlier removal defense, more points are removed as the number of standard deviations from the average decreases.}
    \label{fig:defenses}
\end{figure}

\subsubsection{Stronger point removal defenses.} In Figure~\ref{fig:defenses}, as we increase the number of points removed through the outlier removal and salient point removal defenses for stronger attacks, we see that our three shape attacks are able to maintain their high success rates, compared to the distributional attacks. Therefore, strong point removal defenses are relatively ineffective against our shape attacks.
In practice, removing such a large amount of points is not recommended, as we see that removing nearly half of the points in each benign point cloud also causes PointNet to misclassify more than 20\% of the benign point clouds.

\pgfplotsset{
    cycle list/Dark2,
    cycle multiindex* list={
        mark list*\nextlist
        Dark2\nextlist
    }
}
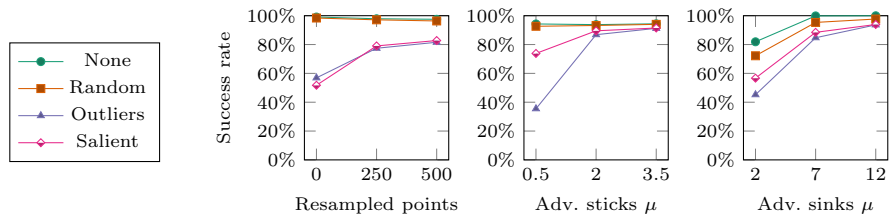
\begin{figure}[tp]
    \centering
    \hspace{7.3em}
    \begin{tikzpicture}
    \begin{axis}[
        name=left,
        xlabel=\scriptsize{Resampled points},
        ylabel=\scriptsize{Success rate},
        ymin=0,
        ymax=1,
        width=3.5cm,
        height=3.5cm,
        legend style={
            overlay,
            at={(-1,0.4)},
            anchor=east,
            font=\scriptsize},
        yticklabel={\scriptsize\pgfmathparse{\tick*100}\pgfmathprintnumber{\pgfmathresult}\%},
        xtick={0,250,500},
        xticklabels={\scriptsize{0},\scriptsize{250},\scriptsize{500}}
    ]
        \addplot+[mark size=1.5pt] table [x=resample,y=none,col sep=comma] {perturb_resample_blend.csv};
        \addplot+[mark size=1.5pt] table [x=resample,y=rand_remove,col sep=comma] {perturb_resample_blend.csv};
        \addplot+[mark size=1.5pt] table [x=resample,y=outliers,col sep=comma] {perturb_resample_blend.csv};
        \addplot+[mark size=1.5pt] table [x=resample,y=salient,col sep=comma] {perturb_resample_blend.csv};
        \legend{None,Random,Outliers,Salient}
    \end{axis}
    \begin{axis}[
        name=right,
        xlabel=\scriptsize{Adv. sticks $\mu$},
        ymin=0,
        ymax=1,
        width=3.5cm,
        height=3.5cm,
        at={($(left.north east)+(1cm,0)$)},
        anchor=north west,
        xtick={0.5,2,3.5},
        yticklabel={\scriptsize\pgfmathparse{\tick*100}\pgfmathprintnumber{\pgfmathresult}\%},
        xticklabel={\scriptsize\pgfmathprintnumber{\tick}}
    ]
        \addplot+[mark size=1.5pt] table [x=resample,y=none,col sep=comma] {adv_sticks_blend.csv};
        \addplot+[mark size=1.5pt] table [x=resample,y=rand_remove,col sep=comma] {adv_sticks_blend.csv};
        \addplot+[mark size=1.5pt] table [x=resample,y=outliers,col sep=comma] {adv_sticks_blend.csv};
        \addplot+[mark size=1.5pt] table [x=resample,y=salient,col sep=comma] {adv_sticks_blend.csv};
    \end{axis}
    \begin{axis}[
        name=right2,
        xlabel=\scriptsize{Adv. sinks $\mu$},
        ymin=0,
        ymax=1,
        width=3.5cm,
        height=3.5cm,
        at={($(right.north east)+(1cm,0)$)},
        anchor=north west,
        xtick={2,7,12},
        yticklabel={\scriptsize\pgfmathparse{\tick*100}\pgfmathprintnumber{\pgfmathresult}\%},
        xticklabel={\scriptsize\pgfmathprintnumber{\tick}}
    ]
        \addplot+[mark size=1.5pt] table [x=resample,y=none,col sep=comma] {adv_sinks_blend.csv};
        \addplot+[mark size=1.5pt] table [x=resample,y=rand_remove,col sep=comma] {adv_sinks_blend.csv};
        \addplot+[mark size=1.5pt] table [x=resample,y=outliers,col sep=comma] {adv_sinks_blend.csv};
        \addplot+[mark size=1.5pt] table [x=resample,y=salient,col sep=comma] {adv_sinks_blend.csv};
    \end{axis}
    \end{tikzpicture}
    
    \caption{The success rates of the perturbation resampling (left), adversarial sticks (middle), and adversarial sinks (right) attacks against different point removal defenses on the PointNet architecture as the number of resampled/perturbed points changes. Essentially, we are ``blending'' between a distributional attack and a shape attack to show the effect of resampling or perturbing more points on the success rates.}
    \label{fig:blend_attack}
\end{figure}

\subsubsection{Blending between a shape attack and a distributional attack.} In Figure~\ref{fig:blend_attack}, we show the effect of changing the number of sampled/influenced points from the shape attacks. This is essentially blending between a shape attack and a simple ``distributional attack''. We see that the attacks rely heavily on techniques like resampling to be robust against point removal defenses. Additionally, we see that resampling does not significantly affect the success rate of an attack against an undefended PointNet.

\begin{table*}[tp]
    \centering
    \caption{The success rates of some adversarial attacks with different $\epsilon$ values (magnitude of perturbations) against different defenses on the PointNet and PointNet++ architectures. The highest attack success rate for each model and defense is bolded.}
    
    \begin{adjustbox}{width=\textwidth,center}
    \setlength{\tabcolsep}{1em}
    {\renewcommand{\arraystretch}{1.2}%
    \begin{tabular}{lc|cccc|cccc}
        \hline
        & & \multicolumn{4}{c|}{PointNet} & \multicolumn{4}{c}{PointNet++}\\
        Attacks & $\epsilon$ & None & Random & Outliers & Salient & None & Random & Outliers & Salient\\
        \hline
        \multirow{3}{*}{Iter. gradient $L_2$} & 1 & 96.9\% & 94.5\% & 46.6\% & 41.5\% & 99.7\% & 80.1\% & 30.5\% & 28.1\%\\
        & 2 & 99.1\% & 98.4\% & 56.9\% & 51.8\% & \textbf{100.0\%} & 96.2\% & 44.4\% & 46.6\%\\
        & 3 & \textbf{99.4\%} & \textbf{99.3\%} & 62.1\% & 55.8\% & \textbf{100.0\%} & \textbf{98.7\%} & 53.0\% & 55.5\%\\
        \hline
        \multirow{3}{*}{Perturb. resample} & 1 & 93.2\% & 91.0\% & 74.0\% & 74.7\% & 92.8\% & 87.2\% & 78.4\% & 71.3\%\\
        & 2 & 97.5\% & 96.4\% & 81.6\% & \textbf{82.8\%} & 98.4\% & 95.2\% & 86.7\% & 81.0\%\\
        & 3 & 98.7\% & 98.0\% & \textbf{84.3\%} & 84.3\% & 99.6\% & 98.1\% & \textbf{92.3\%} & \textbf{88.3\%}\\
        \hline
    \end{tabular}}
    \end{adjustbox}
    
    \label{table:attack_eps}
\end{table*}

\begin{table*}[tp]
    \centering
    \caption{The success rate of the iterative gradient $L_2$ attack with random dropout of 524 (out of 1024) points in each of the 100 iterations.}
    
    \begin{adjustbox}{width=0.6\textwidth,center}
    \setlength{\tabcolsep}{1em}
    {\renewcommand{\arraystretch}{1.2}%
    \begin{tabular}{l|cccc}
        \hline
        & \multicolumn{4}{c}{Defense}\\
        Models & None & Random & Outliers & Salient\\
        \hline
        PointNet & 98.6\% & 98.5\% & 70.1\% & 78.8\%\\
        PointNet++ & 59.6\% & 63.2\% & 40.4\% & 45.4\%\\
        DGCNN & 15.9\% & 24.0\% & 15.9\% & 23.4\%\\
        \hline
    \end{tabular}}
    \end{adjustbox}
    
    \label{table:dropout_attack}
\end{table*}

\subsubsection{Using different $\epsilon$ values in the attacks.} For some attacks that requires an $\epsilon$ hyperparameter, we show the effect of changing the $\epsilon$ values on the success rates of the attacks in Table~\ref{table:attack_eps}. As expected, the success rate of each attack increases as $\epsilon$ increases, which exemplifies the tradeoff between perceptibility and attack success rate.

\subsubsection{Random dropout for iterative gradient $L_2$.} A possible alternative avenue of exploration for attacks that are robust against point-removal defenses is to try to apply those defenses while attacks. To test this alternative, we randomly remove around half of the points in each iteration of iterative gradient $L_2$, and report our results in Table~\ref{table:dropout_attack}. This should help spread out perturbations across more points, so removing some would not lead to the attack failing. Although this does lead to some improvement over vanilla iterative gradient $L_2$ against a defended PointNet, it is not effective against other architectures.

\subsubsection{Perceptibility vs robustness.} There is a definite trade off between the visual perceptibility of the attacks and their robustness against defenses. The shape and distributional attacks represent different extremes of this spectrum. In practice, we think that the shape attacks can be used as \textit{physical} attacks on real-world objects \textit{before} scanning, and distributional attacks can be used as imperceptible \textit{digital} attacks. We believe that effective, but imperceptible physical attacks are very difficult to craft without creating large, focused perturbations in certain areas of an object, as the adversarial features generated by an attack will be easily removed or ignored through the defenses. Small perturbations spread out across an entire shape are too easily masked through point dropout.

\section{Conclusion}
We examine previous distributional attacks, which optimize for human imperceptibility, and shape attacks, which are more realistic. We also introduce three new shape attacks for 3D point cloud classifiers, which are robust against previous defenses or preprocessing steps. These attacks show that changing the shape of a 3D object or adding a few adversarial features are viable options for generating effective adversarial attacks. Our work is an important step in understanding the vulnerability of point cloud neural networks and defenses against adversarial attacks. For future work, it is important to create defenses that are robust to our shape attacks, and also examine the effectiveness of these shape attacks in the context of other tasks, like point cloud segmentation.

\section*{Acknowledgements}
This work was supported in part by NSF awards CNS-1730158, ACI-1540112, ACI-1541349, OAC-1826967, the University of California Office of the President, and the University of California San Diego’s California Institute for Telecommunications and Information Technology/Qualcomm Institute. Thanks to CENIC for the 100Gpbs networks. We want to thank Battista Biggio from the University of Caliagri for feedback on a draft of this manuscript.

\bibliographystyle{splncs04}
\bibliography{main}

\begin{thebibliography}{10}
\providecommand{\url}[1]{\texttt{#1}}
\providecommand{\urlprefix}{URL }
\providecommand{\doi}[1]{https://doi.org/#1}

\bibitem{bernardini1999ball}
Bernardini, F., Mittleman, J., Rushmeier, H., Silva, C., Taubin, G.: The
  ball-pivoting algorithm for surface reconstruction. IEEE Transactions on
  Visualization and Computer Graphics  \textbf{5}(4),  349--359 (1999)

\bibitem{biggio2013evasion}
Biggio, B., Corona, I., Maiorca, D., Nelson, B., {\v{S}}rndi{\'c}, N., Laskov,
  P., Giacinto, G., Roli, F.: {Evasion Attacks Against Machine Learning at Test
  Time}. In: Joint {European} Conference on Machine Learning and Knowledge
  Discovery in Databases. pp. 387--402. Springer (2013)

\bibitem{biggio2018wild}
Biggio, B., Roli, F.: {Wild Patterns: Ten Years After the Rise of Adversarial
  Machine Learning}. Pattern Recognition  \textbf{84},  317--331 (2018)

\bibitem{brown2017adversarial}
Brown, T.B., Mané, D., Roy, A., Abadi, M., Gilmer, J.: {Adversarial Patch}.
  arXiv preprint arXiv:1712.09665  (2017)

\bibitem{cao2019adversarial}
Cao, Y., Xiao, C., Yang, D., Fang, J., Yang, R., Liu, M., Li, B.: {Adversarial
  Objects Against LiDAR-Based Autonomous Driving Systems}. arXiv preprint
  arXiv:1907.05418  (2019)

\bibitem{carlini2017towards}
Carlini, N., Wagner, D.: {Towards Evaluating the Robustness of Neural
  Networks}. In: 2017 IEEE Symposium on Security and Privacy. pp. 39--57. IEEE
  (2017)

\bibitem{deng2018ppf}
Deng, H., Birdal, T., Ilic, S.: {PPF-FoldNet: Unsupervised Learning of Rotation
  Invariant 3D Local Descriptors}. arXiv preprint arXiv:1808.10322  (2018)

\bibitem{dong2018boosting}
Dong, Y., Liao, F., Pang, T., Su, H., Zhu, J., Hu, X., Li, J.: {Boosting
  Adversarial Attacks with Momentum}. arXiv preprint  (2018)

\bibitem{edelsbrunner1983shape}
Edelsbrunner, H., Kirkpatrick, D., Seidel, R.: On the shape of a set of points
  in the plane. IEEE Transactions on Information Theory  \textbf{29}(4),
  551--559 (1983)

\bibitem{goodfellow2014explaining}
Goodfellow, I., Shlens, J., Szegedy, C.: {Explaining and Harnessing Adversarial
  Examples}. arXiv preprint arXiv:1412.6572  (2014)

\bibitem{guo2018low}
Guo, C., Frank, J.S., Weinberger, K.Q.: Low frequency adversarial perturbation.
  arXiv preprint arXiv:1809.08758  (2018)

\bibitem{kingma2014adam}
Kingma, D.P., Ba, J.: {Adam: A method for stochastic optimization}. arXiv
  preprint arXiv:1412.6980  (2014)

\bibitem{kurakin2016adversarialphysical}
Kurakin, A., Goodfellow, I., Bengio, S.: {Adversarial Examples in the Physical
  World}. arXiv preprint arXiv:1607.02533  (2016)

\bibitem{kurakin2016adversarialscale}
Kurakin, A., Goodfellow, I., Bengio, S.: {Adversarial Machine Learning at
  Scale}. arXiv preprint arXiv:1611.01236  (2016)

\bibitem{lee1980two}
Lee, D.T., Schachter, B.J.: Two algorithms for constructing a {Delaunay}
  triangulation. International Journal of Computer \& Information Sciences
  \textbf{9}(3),  219--242 (1980)

\bibitem{liu2019extending}
Liu, D., Yu, R., Su, H.: {Extending Adversarial Attacks and Defenses to Deep 3D
  Point Cloud Classifiers}. arXiv preprint arXiv:1901.03006  (2019)

\bibitem{madry2017towards}
Madry, A., Makelov, A., Schmidt, L., Tsipras, D., Vladu, A.: {Towards Deep
  Learning Models Resistant to Adversarial Attacks}. arXiv preprint
  arXiv:1706.06083  (2017)

\bibitem{moosavi2017universal}
Moosavi-Dezfooli, S.M., Fawzi, A., Fawzi, O., Frossard, P.: Universal
  adversarial perturbations. In: Proceedings of the IEEE Conference on Computer
  Vision and Pattern Recognition. pp. 1765--1773 (2017)

\bibitem{moosavi2016deepfool}
Moosavi-Dezfooli, S.M., Fawzi, A., Frossard, P.: {DeepFool: A simple and
  accurate method to fool deep neural networks}. In: Proceedings of the IEEE
  Conference on Computer Vision and Pattern Recognition. pp. 2574--2582 (2016)

\bibitem{papernot2016limitations}
Papernot, N., McDaniel, P., Jha, S., Fredrikson, M., Celik, Z.B., Swami, A.:
  {The Limitations of Deep Learning in Adversarial Settings}. In: 2016 IEEE
  European Symposium on Security and Privacy (EuroS\&P). pp. 372--387. IEEE
  (2016)

\bibitem{papernot2016distillation}
Papernot, N., McDaniel, P., Wu, X., Jha, S., Swami, A.: {Distillation as a
  Defense to Adversarial Perturbations Against Deep Neural Networks}. In: 2016
  IEEE Symposium on Security and Privacy (SP). pp. 582--597. IEEE (2016)

\bibitem{qi2017frustum}
Qi, C.R., Liu, W., Wu, C., Su, H., Guibas, L.J.: {Frustum PointNets for 3D
  Object Detection from RGB-D Data}. arXiv preprint arXiv:1711.08488  (2017)

\bibitem{qi2017pointnet}
Qi, C.R., Su, H., Mo, K., Guibas, L.J.: {PointNet: Deep Learning on Point Sets
  for 3D Classification and Segmentation}. Proceedings of the IEEE Conference
  on Computer Vision and Pattern Recognition  \textbf{1}(2), ~4 (2017)

\bibitem{qi2017pointnetplusplus}
Qi, C.R., Yi, L., Su, H., Guibas, L.J.: {PointNet++: Deep Hierarchical Feature
  Learning on Point Sets in a Metric Space}. In: Advances in Neural Information
  Processing Systems. pp. 5099--5108 (2017)

\bibitem{szegedy2013intriguing}
Szegedy, C., Zaremba, W., Sutskever, I., Bruna, J., Erhan, D., Goodfellow, I.,
  Fergus, R.: Intriguing properties of neural networks. arXiv preprint
  arXiv:1312.6199  (2013)

\bibitem{tsai2020robust}
Tsai, T., Yang, K., Ho, T.Y., Jin, Y.: Robust adversarial objects against deep
  learning models. In: Proceedings of the {AAAI Conference on Artificial
  Intelligence}. vol.~34, pp. 954--962 (2020)

\bibitem{wang2017cnn}
Wang, P.S., Liu, Y., Guo, Y.X., Sun, C.Y., Tong, X.: {O-CNN: Octree-based
  convolutional neural networks for 3D shape analysis}. ACM Transactions on
  Graphics  \textbf{36}(4), ~72 (2017)

\bibitem{dgcnn}
Wang, Y., Sun, Y., Liu, Z., Sarma, S.E., Bronstein, M.M., Solomon, J.M.:
  Dynamic graph {CNN} for learning on point clouds. {ACM Transactions on
  Graphics (TOG)}  (2019)

\bibitem{wicker2019robustness}
Wicker, M., Kwiatkowska, M.: {Robustness of 3D Deep Learning in an Adversarial
  Setting}. In: Proceedings of the IEEE Conference on Computer Vision and
  Pattern Recognition. pp. 11767--11775 (2019)

\bibitem{wong2017provable}
Wong, E., Kolter, J.Z.: {Provable Defenses against Adversarial Examples via the
  Convex Outer Adversarial Polytope}. arXiv preprint arXiv:1711.00851  (2017)

\bibitem{wu2015shapenet}
Wu, Z., Song, S., Khosla, A., Yu, F., Zhang, L., Tang, X., Xiao, J.: {3D
  ShapeNets: A Deep Representation for Volumetric Shapes}. In: Proceedings of
  the IEEE Conference on Computer Vision and Pattern Recognition. pp.
  1912--1920 (2015)

\bibitem{xiang2018generating}
Xiang, C., Qi, C.R., Li, B.: {Generating 3D Adversarial Point Clouds}. arXiv
  preprint arXiv:1809.07016  (2018)

\bibitem{yang2019adversarial}
Yang, J., Zhang, Q., Fang, R., Ni, B., Liu, J., Tian, Q.: {Adversarial Attack
  and Defense on Point Sets}. arXiv preprint arXiv:1902.10899  (2019)

\bibitem{yianilos1993data}
Yianilos, P.N.: {Data Structures and Algorithms for Nearest Neighbor Search in
  General Metric Spaces}. In: Proceedings of the fourth annual {ACM-SIAM}
  symposium on {Discrete} algorithms. vol.~93, pp. 311--21 (1993)

\bibitem{zheng2018learning}
Zheng, T., Chen, C., Ren, K., et~al.: {Learning Saliency Maps for Adversarial
  Point-Cloud Generation}. arXiv preprint arXiv:1812.01687  (2018)

\bibitem{zhou2018deflecting}
Zhou, H., Chen, K., Zhang, W., Fang, H., Zhou, W., Yu, N.: {Deflecting 3D
  Adversarial Point Clouds Through Outlier-Guided Removal}. arXiv preprint
  arXiv:1812.11017  (2018)

\end{thebibliography}

\end{document}